\documentclass[conference]{IEEEtran}
\IEEEoverridecommandlockouts

\usepackage{cite}
\usepackage{amsmath,amssymb,amsfonts}
\usepackage{algorithmic}
\usepackage{graphicx}
\usepackage{textcomp}
\usepackage{xcolor}
\usepackage{multirow}
\usepackage{booktabs}
\usepackage{amsmath}
\usepackage{amssymb}
\def\BibTeX{{\rm B\kern-.05em{\sc i\kern-.025em b}\kern-.08em
    T\kern-.1667em\lower.7ex\hbox{E}\kern-.125emX}}
\begin{document}


\title{PA-Net: Precipitation-Adaptive Mixture-of-Experts for Long-Tail Rainfall Nowcasting}

\author{ {Xinyu Xiao, Sen Lei, Eryun Liu, Shiming Xiang, Hao Li, Cheng Yuan, Yuan Qi, Qizhao Jin}


\IEEEcompsocitemizethanks 
{
\IEEEcompsocthanksitem
(\textit{Corresponding author: Qizhao Jin})
}
\thanks {Xinyu Xiao, Hao Li, Cheng Yuan and Yuan Qi are with Fudan University}

\thanks {Sen Lei is with the School of Information Science and Technology, Southwest Jiaotong University}

\thanks {Eryun Liu is with Zhejiang University}

\thanks {Shiming Xiang is with the Institute of Automation, Chinese Academy of Sciences}

\thanks {Qizhao Jin is with the Public Meteorological Service Center, China Meteorological Administration}

}

\maketitle

\begin{abstract}

Precipitation nowcasting is vital for flood warning, agricultural management, and emergency response, yet two bottlenecks persist: the prohibitive cost of modeling million-scale spatiotemporal tokens from multi-variate atmospheric fields, and the extreme long-tailed rainfall distribution where heavy-to-torrential events—those of greatest societal impact—constitute fewer than 0.1\% of all samples. We propose the \textbf{Precipitation-Adaptive Network (PA-Net)}, a Transformer framework whose computational budget is explicitly governed by rainfall intensity. Its core component, \textbf{Precipitation-Adaptive MoE (PA-MoE)}, dynamically scales the number of activated experts per token according to local precipitation magnitude, channeling richer representational capacity toward the rare yet critical heavy-rainfall tail. A \textbf{Dual-Axis Compressed Latent Attention} mechanism factorizes spatiotemporal attention with convolutional reduction to manage massive context lengths, while an intensity-aware training protocol progressively amplifies learning signals from extreme-rainfall samples. Experiments on ERA5 demonstrate consistent improvements over state-of-the-art baselines, with particularly significant gains in heavy-rain and rainstorm regimes.

\end{abstract}





\maketitle


\section{Introduction}

Short-term rainfall forecasting within a horizon of several hours—commonly referred to as precipitation nowcasting—underpins a broad spectrum of societal applications, from agricultural planning and urban flood management to transportation logistics and disaster preparedness. In practice, delivering simultaneously timely and reliable predictions over extended lead times, particularly for intense convective events, remains an unresolved operational challenge.

For much of the past half-century, numerical weather prediction (NWP) systems \cite{bauer2015quiet} have served as the workhorse of operational rainfall forecasting. By numerically integrating the governing equations of atmospheric fluid dynamics from observed initial conditions, NWP provides physically grounded projections of future atmospheric states. Nevertheless, the computational burden associated with solving high-dimensional partial differential equations over fine spatiotemporal grids \cite{sun2014use} severely constrains the update frequency and spatial resolution attainable in real-time operations, limiting the practical utility of NWP for rapid-refresh nowcasting.

\begin{table}[h]
  \caption{Empirical frequency of precipitation intensity categories across the ERA5 and WeatherBench archives. Categories are ordered by ascending severity: Rainless (RL), Light Rain (LR), Moderate Rain (MR), Heavy Rain (HR), and Rainstorm (RS).}
  	\renewcommand{\arraystretch}{0.7}
	\centering
    \resizebox{\linewidth}{!}
{
\begin{tabular}{c|c|c|c|c|c}
\toprule
{Intensity} & RL   & LR   & MR & HR     & RS          \\ 
Precipitation(mm) & { [}0, 0.1) & {[}0.1, 4.0) & {[}4.0, 13.0) & {[}13.0, 25.0) & {[}25.0, $\infty$) \\ 
             \midrule
ERA5         & 80.76\%      & 18.22\%        & 0.96\%          & 0.06\%           & 0.006\%              \\
WeatherBench & 83.59\%      & 15.94\%       & 0.44\%         & 0.02\%           & 0.003\%              \\ 
\bottomrule
\end{tabular}
}
  \label{tab:data}
\end{table}

A defining characteristic of real-world rainfall data, as quantified in Table~\ref{tab:data}, is its extreme distributional skewness: more than 80\% of grid cells record negligible or zero accumulation, while heavy and torrential events collectively account for fewer than 0.1\% of all observations. This pronounced long-tailed structure constitutes a fundamental obstacle for data-driven models, which inevitably absorb the statistical bias of the training distribution and devote the bulk of their representational capacity to the dominant dry and light-rain majority.

The past decade has witnessed a surge of deep learning approaches that recast nowcasting as a spatiotemporal sequence extrapolation problem \cite{ShiCWYWW15,PrecipitationNowcasting17,gao2024prediff}. Gao et al.\ \cite{0001S0Z00Y22} introduced a space-time efficient Transformer architecture for radar-based rainfall prediction at lead times up to 60 minutes. Ravuri et al.\ \cite{ravuri2021skilful} presented a deep generative framework (DGMR) capable of synthesizing plausible radar echo sequences extending to 90-minute horizons. Although these methods achieve commendable performance on light-to-moderate precipitation regimes, their skill degrades markedly when confronted with tail-end extremes—precisely the events of greatest meteorological and societal consequence. The root cause is structural: uniform allocation of model capacity across all intensity regimes allows the overwhelming volume of low-intensity samples to dominate the optimization landscape, suppressing the network's ability to faithfully represent the rare but high-impact heavy-rainfall patterns \cite{zhang2023deep}. This systematic bias erodes forecast reliability in the very scenarios—severe storms, flash floods, urban inundation—where accuracy matters most \cite{bi2023accurate}.

From a physical standpoint, precipitation genesis involves intricate multi-scale couplings among diverse atmospheric variables spanning both space and time. Convolutional neural networks, constrained by their local receptive fields, are inherently ill-suited to capturing the long-range teleconnections that govern the organization and intensification of rainfall systems. Motivated by the demonstrated efficacy of attention mechanisms in modeling global contextual relationships \cite{DBLP:conf/iclr/ParkK22}, the Transformer architecture has rapidly gained traction in meteorological forecasting. However, when multi-variate atmospheric fields are discretized across high-resolution spatial grids and extended temporal windows, the resulting token count can reach into the millions—far exceeding the practical capacity of standard full-attention Transformers. This scalability bottleneck, compounded by the inability of dense, uniform attention to preferentially encode rare extreme-precipitation signals, limits the applicability of existing Transformer-based nowcasting systems to long-range, high-resolution operational settings.

To jointly address the scalability and distributional imbalance challenges outlined above, this paper introduces the \textbf{Precipitation-Adaptive Network (PA-Net)}, a novel Transformer-based framework whose architectural design is explicitly guided by the physical structure of rainfall intensity distributions. The core contributions of this work are summarized as follows:

\begin{itemize}
    \item \textbf{Precipitation-Adaptive Mixture-of-Experts (PA-MoE).} We propose a new MoE mechanism in which the number of activated experts per token is dynamically governed by the local precipitation intensity rather than by a fixed top-$k$ rule. Tokens situated in heavy-rainfall regions—corresponding to the long tail of the intensity distribution—recruit a large ensemble of specialist sub-networks, granting the model augmented representational power where the prediction task is most demanding. Conversely, tokens in dry or light-rain zones are processed by a compact expert subset, avoiding superfluous computation on well-represented regimes. This intensity-driven allocation explicitly aligns model capacity with the intrinsic difficulty landscape of the nowcasting problem.

    \item \textbf{Dual-Axis Compressed Latent Attention (DACLA).} To overcome the prohibitive quadratic cost of full spatiotemporal self-attention, we factorize the attention computation into sequential intra-frame spatial and cross-frame temporal stages, each augmented with convolutional dimensionality reduction. This decomposition shrinks the effective token count along each axis independently, enabling PA-Net to process million-scale spatiotemporal contexts within practical memory and compute budgets while preserving long-range dependency modeling.

    \item \textbf{Intensity-Aware Training Protocol.} We design a composite optimization strategy comprising (i) a soft Dice prediction loss that inherently counteracts class-frequency bias, (ii) a dual-component PA-MoE regularizer that jointly enforces dispatch equity and routing decisiveness, and (iii) a curriculum-style intensity re-weighting schedule that progressively amplifies the supervisory signal from extreme-rainfall samples as training progresses. Together, these elements ensure that the network devotes sustained learning effort to the meteorologically critical tail of the precipitation distribution without sacrificing baseline performance on common rainfall regimes.
\end{itemize}


Extensive experiments on the ERA5 benchmark demonstrate that PA-Net achieves substantial improvements over state-of-the-art baselines across all precipitation intensity categories, with particularly pronounced gains in the heavy-rain and rainstorm regimes. Ablation studies confirm that each proposed component contributes meaningfully to overall forecast skill, validating the effectiveness of coupling expert allocation to precipitation magnitude as a principled strategy for long-tail meteorological prediction.

\section{Related Work}



\subsection{Data-Driven Precipitation Nowcasting}

Deep learning has emerged as a compelling paradigm for precipitation nowcasting, with the dominant line of research casting the problem as spatiotemporal sequence extrapolation from radar observations. The seminal ConvLSTM~\cite{ShiCWYWW15} replaced fully connected gates with convolutional operators to jointly model spatial and temporal patterns in radar echo imagery. This design spawned a rich family of recurrent variants—including TrajGRU~\cite{PrecipitationNowcasting17}, MotionRNN~\cite{DBLP:conf/cvpr/WuY0L21}, and PredRNN~\cite{DBLP:conf/nips/WangLWGY17}—that progressively enhanced the modeling of spatiotemporal dynamics. Beyond the recurrent paradigm, subsequent efforts have explored CNN-based~\cite{DBLP:conf/cvpr/WuY0L21} and Transformer-based~\cite{DBLP:conf/nips/0001S0Z00Y22} backbones to capture long-range spatial dependencies more effectively. Meanwhile, the rapid maturation of generative modeling has given rise to radar-conditioned probabilistic forecasters such as DGMR~\cite{ravuri2021skilful}, which draws multiple plausible future radar sequences from the conditional distribution given historical observations.

Despite their promising overall skill, radar-only models inherently lack access to the broader atmospheric state variables that govern the genesis and intensification of precipitation systems, making it particularly difficult to represent extreme rainfall events. NowcastNet~\cite{nowcastnet} addresses this gap by coupling physical evolution operators with neural networks, yet its architecture remains tightly bound to specific physical parameterizations. SCCN~\cite{DBLP:conf/icdm/XiaoJMXP22} takes an alternative route, exploiting spatiotemporal consistency constraints to align tail-distributed features across multi-modal meteorological inputs; however, its effectiveness deteriorates at longer lead times as the consistency assumption weakens.

Our work departs from these approaches in two fundamental respects. First, rather than treating all precipitation regimes with uniform model capacity, PA-Net introduces a Precipitation-Adaptive MoE mechanism that dynamically scales expert allocation in proportion to local rainfall intensity, explicitly channeling richer representational resources toward the long-tail heavy-precipitation events that prior methods systematically under-serve. Second, PA-Net leverages multi-variate atmospheric fields to capture the inter-variable couplings that drive convective organization, while the Dual-Axis Compressed Latent Attention enables efficient processing of the resulting large-scale spatiotemporal token sequences without sacrificing long-range dependency modeling.

\subsection{Tokenization in Deep Learning}

Tokenization—the conversion of raw data into discrete representational units amenable to neural processing—constitutes a foundational preprocessing step across both natural language processing (NLP) and computer vision (CV).

In NLP, tokenization strategies have progressed from rudimentary whitespace- and punctuation-based segmentation to sophisticated subword algorithms. Byte Pair Encoding (BPE)~\cite{sennrich2015neural} marked a pivotal advance by enabling graceful handling of rare and morphologically complex words through iterative merging of frequent character pairs. SentencePiece~\cite{kudo2018sentencepiece} subsequently generalized subword segmentation to language-agnostic raw text, eliminating the need for hand-crafted preprocessing pipelines. The widespread adoption of WordPiece tokenization in Transformer-based language models such as BERT~\cite{kenton2019bert} further demonstrated the efficiency of subword decomposition for managing large and diverse vocabularies.

In CV, the Vision Transformer (ViT)~\cite{DBLP:conf/iclr/DosovitskiyB0WZ21} pioneered patch-based tokenization by partitioning an image into fixed-size spatial tiles and projecting each tile into the Transformer's embedding space, thereby enabling global context aggregation via self-attention. To alleviate the quadratic computational burden associated with high-resolution inputs, Swin Transformer~\cite{liu2021Swin,liu2021swinv2} introduced a hierarchical shifted-window tokenization scheme that restricts attention to local neighborhoods while permitting cross-window information exchange. Discrete tokenization offers yet another perspective: DALL·E~\cite{ramesh2021zero} employs a VQ-VAE~\cite{van2017neural} codebook to compress images into a sequence of discrete codes structurally analogous to text tokens, facilitating unified cross-modal generation.

PA-Net adopts a meteorologically informed tokenization scheme: spatial patches are augmented with solar-geometric positional encodings that embed physically meaningful geotemporal information, and the subsequent PA-MoE routing layer reorganizes token processing not by position but by precipitation intensity, ensuring that the computational graph reflects the intrinsic complexity landscape of the atmospheric prediction task rather than purely geometric locality.

\section{Background}

\subsection{Task Definition}

Short-range precipitation prediction (nowcasting) seeks to extrapolate forthcoming rainfall fields from a collection of recent and current atmospheric observations. These observations encompass multiple meteorological variables—such as accumulated rainfall, relative humidity, surface temperature, and horizontal wind components—organized as a gridded field $\mathbf{M}_t \in \mathbb{R}^{h \times w \times l}$ at each discrete time index $t$, where the spatial domain is tessellated into an $h \times w$ lattice and $l$ indexes the variable channels. Provided with $s$ consecutive historical snapshots up to and including the present moment, the nowcasting objective amounts to recovering the most likely future precipitation sequence:
\begin{small}
\begin{equation}
    \label{E1}
    \hat{\mathbf{R}}_{t+1}, \hat{\mathbf{R}}_{t+2}, \ldots, \hat{\mathbf{R}}_{t+j}= \mathcal{F}(\mathbf{M}_{t-s+1}, \mathbf{M}_{t-s+2}, \ldots, \mathbf{M}_{t}),
\end{equation}
\end{small}\noindent
where $\mathbf{R}_{t+1} \in \mathbb{N}^{h \times w}$ encodes the rainfall accumulation between time steps $t$ and $t+1$ over the target domain, $j$ specifies the forecast lead-time horizon, and $s$ is the look-back window length. This mapping relies exclusively on directly observed atmospheric states to infer unobserved future rainfall, which constitutes the canonical formulation in contemporary precipitation modeling.

\subsection{Open Difficulties in Precipitation Nowcasting}

Data-driven meteorological prediction has witnessed remarkable strides in recent years. Transformer-based architectures, in particular, enable the joint assimilation of heterogeneous atmospheric variables within a unified end-to-end pipeline, pushing neural forecasting skill toward parity with operational numerical weather prediction (NWP) systems. Nonetheless, two fundamental bottlenecks persist.

\textit{First}, when diverse meteorological fields are fused with multi-step temporal sequences, the resulting spatiotemporal token count can escalate to millions—far exceeding the effective context window of even the most capable contemporary large-scale models. This combinatorial explosion in sequence length constitutes a critical scalability barrier for next-generation weather prediction networks.

\textit{Second}, the highly skewed, long-tailed nature of rainfall intensity distributions poses a severe modeling challenge. Heavy-to-extreme precipitation events are statistically rare yet disproportionately consequential. Existing architectures uniformly allocate model capacity across all intensity regimes, leading to systematic under-representation of high-impact tail events. Consequently, current nowcasting systems frequently underestimate the magnitude and spatial footprint of intense convective rainfall, eroding their operational utility for severe-weather warning applications.

\section{Preliminaries}

\subsection{Transformer Architecture}

Originally put forward by Vaswani et al., the Transformer has become the dominant backbone for sequence-to-sequence tasks spanning natural language understanding, visual recognition, and cross-modal reasoning. Its three pillars are: the self-attention operator, the position-wise feed-forward sub-layer, and positional encoding. The scaled dot-product attention and its multi-head extension are expressed as:
\begin{align}
\text{Attn}(Q, K, V) &= \text{Softmax}\!\left(\frac{QK^{\top}}{\sqrt{d_k}}\right)V, \\
\text{MH}(Q, K, V) &= [\text{head}_1;\;\ldots;\;\text{head}_h]\,W^O,
\end{align}
where $\text{head}_i = \text{Attn}(QW_i^Q, KW_i^K, VW_i^V)$. Each layer further contains a two-layer fully connected sub-network applied element-wise to every position:
\begin{equation}
\text{FFN}(x) = \text{ReLU}(xW_1 + b_1)\,W_2 + b_2.
\end{equation}
Because the architecture possesses no intrinsic notion of order, sinusoidal positional signals are injected to supply sequential structure:
\begin{align}
\text{PE}_{(\text{pos},2i)} &= \sin\!\left(\frac{\text{pos}}{10000^{2i/d_{\text{model}}}}\right), \\
\text{PE}_{(\text{pos},2i+1)} &= \cos\!\left(\frac{\text{pos}}{10000^{2i/d_{\text{model}}}}\right).
\end{align}
Residual shortcuts combined with layer normalization wrap every sub-layer to stabilize gradient flow and accelerate convergence.

\subsection{Mixture-of-Experts Paradigm}

The Mixture-of-Experts (MoE) concept, tracing back to Jacobs et al.\ and later scaled by Shazeer et al.\ through GShard and Switch Transformer, has been substantially refined by the DeepSeekMoE line of work, which partitions experts into \emph{shared} and \emph{routed} pools within a single layer. Shared experts internalize broadly applicable representations, whereas routed experts develop narrow specializations for less frequent patterns. The full expert inventory is written as:
\begin{equation}
\mathcal{E} = \underbrace{\{E_{\text{sh}}^1, \ldots, E_{\text{sh}}^{N_s}\}}_{\text{Shared Pool}} \;\cup\; \underbrace{\{E_{\text{rt}}^1, \ldots, E_{\text{rt}}^{N_r}\}}_{\text{Routed Pool}},
\end{equation}
where $N_s$ and $N_r$ are the pool sizes. The composite layer output fuses both pools:
\begin{equation}
\begin{split}
\text{Out} = \text{LN}\Bigl(x &+ \textstyle\sum_{i \in \mathcal{A}_{\text{sh}}} g_i^{\text{sh}} \cdot E_{\text{sh}}^i(x) \\
&+ \textstyle\sum_{j \in \mathcal{A}_{\text{rt}}} g_j^{\text{rt}} \cdot E_{\text{rt}}^j(x)\Bigr),
\end{split}
\end{equation}
where $\mathcal{A}_{\text{sh}}$ and $\mathcal{A}_{\text{rt}}$ denote the activated index sets, and $g^{\text{sh}}, g^{\text{rt}}$ are the corresponding gating weights. An auxiliary regularizer encourages balanced token dispatch:
\begin{equation}
\mathcal{L}_{\text{bal}} = \lambda \sum_{i=1}^{N} f_i \cdot \bar{g}_i + \mu \cdot \text{Var}(f_1, \ldots, f_N),
\end{equation}
with $f_i$ the fraction of tokens routed to expert $i$ and $\bar{g}_i$ the mean gate value.

\section{METHODOLOGY}

\begin{figure*}[t!]
    \centering
    \includegraphics[width=410pt]{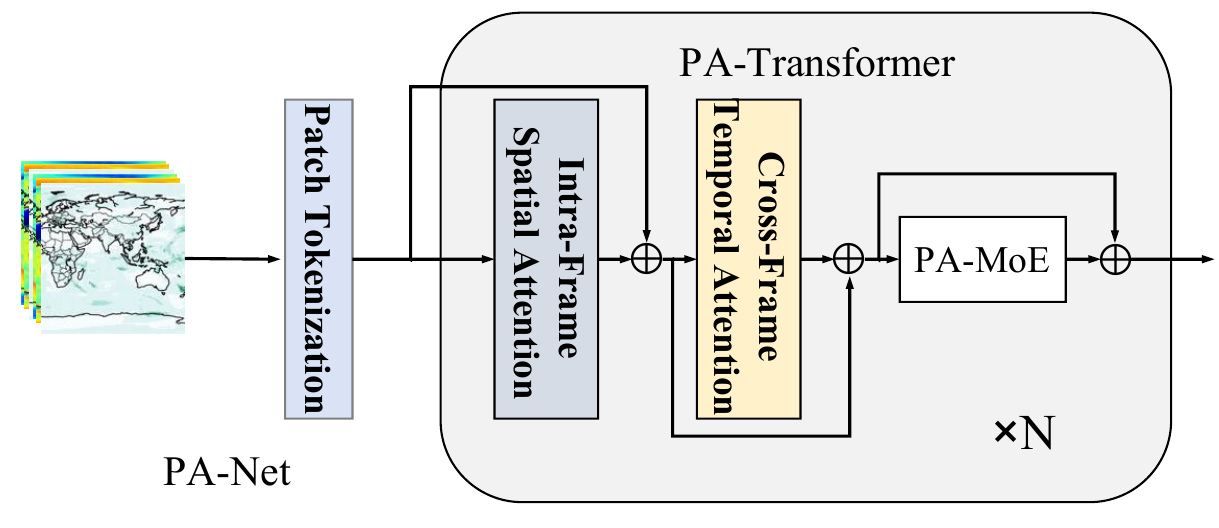}
    \caption{Architecture of the Precipitation-Adaptive Network (PA-Net): a Transformer backbone equipped with Dual-Axis Compressed Latent Attention (DACLA) for scalable spatiotemporal modeling and Precipitation-Adaptive MoE (PA-MoE) for rainfall-intensity-driven expert allocation, jointly enabling enhanced nowcasting skill from light drizzle through torrential downpours.}
    \label{fig:model}
\end{figure*}

In Figure \ref{fig:model}, we present a rainfall-intensity-aware Transformer framework—termed the \textbf{Precipitation-Adaptive Network (PA-Net)}—tailored for short-range precipitation nowcasting over multi-variate meteorological fields. PA-Net rests on three synergistic pillars: (i) a \emph{Precipitation-Adaptive MoE} (PA-MoE) mechanism that dynamically scales the active expert count in proportion to local rainfall intensity, thereby channeling richer model capacity toward the statistically rare yet critically important heavy-precipitation regime; (ii) a \emph{progressive refinement training protocol} that iteratively sharpens predictions for tail-end events; and (iii) a \emph{self-adaptive loss formulation} that re-weights supervisory signals according to precipitation severity. Collectively, these components enable PA-Net to break the inherent bias toward the dominant dry/light-rain majority and substantially improve forecast skill for extreme rainfall.

At a conceptual level, the key departure from prior work is the explicit coupling between precipitation magnitude and computational budget. Rainfall intensity distributions are notoriously long-tailed: the overwhelming majority of grid cells record negligible or zero precipitation, while a tiny fraction experiences torrential downpours. Conventional architectures spread representational resources uniformly across all tokens, implicitly under-serving the tail. PA-MoE rectifies this imbalance by treating rainfall intensity as a first-class routing signal: regions of vigorous convection mobilize a large expert ensemble, moderate-rain zones engage a moderate cohort, and dry regions are processed economically by a minimal set.

To formalize this, we augment the canonical nowcasting objective (Equation~\ref{E1}) by additionally conditioning on high-confidence precipitation priors $\mathbf{R}^{\text{prior}}$ and an intensity-derived routing map $\boldsymbol{\Omega}$:
\begin{small}
\begin{equation}
    \label{E11}
    \begin{split}
        \hat{\mathbf{R}}_{t+1}, \ldots, \hat{\mathbf{R}}_{t+j} = \mathcal{F}\bigl(&\mathbf{M}_{t-s+1}, \ldots, \mathbf{M}_{t}, \\
        &\mathbf{R}^{\text{prior}}_{t+1}, \ldots, \mathbf{R}^{\text{prior}}_{t+j},\; \boldsymbol{\Omega}\bigr),
    \end{split}
\end{equation}
\end{small}
where $\boldsymbol{\Omega}$ encapsulates the per-token expert allocation schedule derived from precipitation intensity. This formulation extends the standard input–output paradigm by injecting physically grounded routing information that steers internal computation toward the most challenging forecast targets.

The remainder of this section details each architectural component and the associated training objectives.

\subsection{PA-Net Backbone Architecture}

\subsubsection{Patch Tokenization}

Following the Vision Transformer (ViT) paradigm, each meteorological snapshot is first decomposed into non-overlapping spatial patches and projected into a latent embedding space. Formally: $\mathbf{Z}_t^p = \mathcal{T}(\mathbf{M}_t)$, where $\mathcal{T}$ comprises a linear projection followed by additive positional embeddings. The resulting patch tensor $\mathbf{Z}_t^p \in \mathbb{R}^{\frac{h}{p} \times \frac{w}{p} \times d}$ maps each $(p \times p)$ spatial tile into a $d$-dimensional token. For notational convenience we write $n = \frac{h}{p} \times \frac{w}{p}$ for the total number of spatial tokens per frame.

\subsubsection{Solar-Geometric Positional Encoding}

To imbue the patch tokens with geotemporal awareness, we encode the solar elevation angle $\alpha_t$ at every grid point through a learnable linear mapping. The solar elevation angle—defined as the angle subtended between the incoming solar ray and the local horizontal plane—is computed via:
\begin{small}
\begin{equation}
    \label{E4}
    \alpha_t = \arccos\!\bigl(\sin\Phi\,\sin\delta + \cos\Phi\,\cos\delta\,\cos\omega_t\bigr),
\end{equation}
\end{small}\noindent
where $\omega_t$ is the local hour angle, $\Phi$ the geographic latitude, and $\delta$ the solar declination. This quantity simultaneously encodes longitude, latitude, and time of day in a single physically meaningful scalar. Moreover, because solar irradiance constitutes the primary thermodynamic driver of atmospheric circulation, $\alpha_t$ furnishes a compact yet informative spatiotemporal fingerprint.

\subsubsection{Dual-Axis Compressed Latent Attention}

The backbone of PA-Net remains grounded in the Transformer paradigm, but replaces monolithic spatiotemporal self-attention with a factorized, compression-augmented variant—\emph{Dual-Axis Compressed Latent Attention} (DACLA)—to manage the prohibitive token count arising from high-resolution multi-step meteorological grids. DACLA decomposes the joint spatiotemporal attention into two sequential stages—\emph{Intra-Frame Spatial Attention} and \emph{Cross-Frame Temporal Attention}—each equipped with convolutional dimensionality reduction. This factorization reduces complexity from quadratic in the full token count to quadratic in each axis independently, while preserving expressive long-range coupling along both dimensions.

{\bfseries Intra-Frame Spatial Attention.}
Within each temporal frame, spatial interactions are modeled after projecting the feature map to a coarser resolution. Given the reshaped input $\tilde{X}_{s} \in \mathbb{R}^{(B \cdot T) \times H \times W \times D}$, a strided 2-D convolution yields:
\begin{equation}
\tilde{X}_{s}^{\downarrow} = \text{SConv}_{2\text{D}}(\tilde{X}_{s}) \in \mathbb{R}^{(B \cdot T) \times H_c \times W_c \times D},
\end{equation}
where $\text{SConv}_{2\text{D}}$ denotes the spatial down-projection kernel. Queries, keys, and values are then derived from this compressed representation and fed into the latent attention operator:
\begin{equation}
A_s = \text{LatentAttn}_{s}(Q_s, K_s, V_s),
\end{equation}
with $Q_s, K_s, V_s$ obtained via layer-normalized linear projections. The attended features are subsequently restored to the native grid through bilinear upsampling:
\begin{equation}
A_s^{\uparrow} = \text{Upsample}_{\text{bilinear}}\!\bigl(A_s,\;(H, W)\bigr).
\end{equation}

{\bfseries Cross-Frame Temporal Attention.}
Temporal coherence across forecast steps is captured by reindexing the spatial-attention output as $\tilde{X}_{t} \in \mathbb{R}^{(B \cdot H \cdot W) \times T \times D}$ and applying a 1-D causal convolution along the time axis:
\begin{equation}
\tilde{X}_{t}^{\downarrow} = \text{TConv}_{1\text{D}}(\tilde{X}_{t}) \in \mathbb{R}^{(B \cdot H \cdot W) \times T_c \times D},
\end{equation}
where $\text{TConv}_{1\text{D}}$ is the temporal down-projection filter. Latent attention is then applied in the compressed temporal domain:
\begin{equation}
A_t = \text{LatentAttn}_{t}(Q_t, K_t, V_t),
\end{equation}
followed by linear interpolation to recover the original temporal resolution:
\begin{equation}
A_t^{\uparrow} = \text{Upsample}_{\text{linear}}\!\bigl(A_t,\;T\bigr).
\end{equation}

\begin{figure}[t!]
    \centering
    \includegraphics[width=240pt]{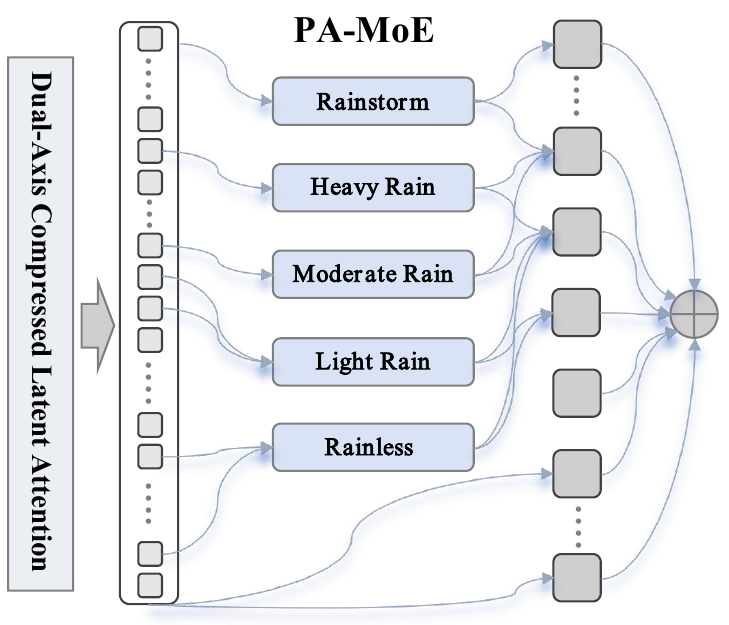}
    \caption{Overview of the Precipitation-Adaptive MoE (PA-MoE): tokens associated with intense rainfall activate a larger expert ensemble, while rainless and light rain are routed to fewer experts, thereby concentrating representational capacity on the rare yet high-impact tail of the precipitation distribution.}
    \label{fig:moe}
\end{figure}

\subsubsection{Precipitation-Adaptive MoE}

Rainfall intensity distributions in real-world datasets are strikingly imbalanced: dry and drizzle conditions account for the vast majority of samples, whereas heavy convective episodes occupy a slender tail that nonetheless dominates societal risk. Conventional MoE layers, which activate a fixed number of experts irrespective of input characteristics, inevitably allocate model capacity in proportion to sample frequency rather than sample importance. This frequency-driven allocation systematically starves the rare extreme-rainfall regime of representational resources.

We introduce the \textbf{Precipitation-Adaptive MoE (PA-MoE)} to remedy this structural mismatch, which can be seen in Figure \ref{fig:moe}. PA-MoE treats local rainfall intensity as an \emph{a priori} complexity indicator and modulates the expert activation budget accordingly: grid cells experiencing intense precipitation recruit a broad coalition of specialist sub-networks, moderate-rainfall cells engage an intermediate cohort, and dry or trace-rain cells are served by a lean subset. This design philosophy ensures that the network's expressive power is concentrated on the long-tail events that are hardest to learn yet most consequential to forecast.

Let $X \in \mathbb{R}^{B \times T \times H \times W \times D}$ denote the latent representation emerging from the DACLA block, and let $\mathbf{I}_r \in \mathbb{R}^{B \times T \times H \times W}$ be the rainfall intensity field (expressed in mm/h) associated with each spatiotemporal token. Two climatologically informed intensity boundaries, $\lambda_{\text{str}}$ (strong-rain threshold) and $\lambda_{\text{wk}}$ (weak-rain threshold), segment the precipitation continuum into three physically distinct regimes. The per-token expert budget is then prescribed as:

\begin{equation}
k_{\text{token}} = 
\begin{cases} 
k_{\text{max}} & \text{if } r \geq \lambda_{\text{str}}, \\[4pt]
k_{\text{med}} & \text{if } \lambda_{\text{wk}} \leq r < \lambda_{\text{str}}, \\[4pt]
k_{\text{min}} & \text{if } r < \lambda_{\text{wk}},
\end{cases}
\end{equation}

\noindent where $r$ is the precipitation rate at the corresponding grid cell, and the integer hyperparameters satisfy $k_{\text{max}} > k_{\text{med}} > k_{\text{min}} \geq 1$. Intuitively, $k_{\text{max}}$ provisions a rich expert ensemble for torrential-rain tokens residing in the distribution tail, $k_{\text{med}}$ supplies a moderate panel for transitional rainfall intensities, and $k_{\text{min}}$ assigns a parsimonious group for the abundant dry or trace-rain tokens. The thresholds $\lambda_{\text{str}}$ and $\lambda_{\text{wk}}$ are derived from the empirical cumulative distribution function (CDF) of the training set—e.g., the 95th and 75th percentiles—so that the allocation naturally adapts to regional climatology.

Given an individual token $x_j$ with assigned budget $k_j$, the routing procedure first evaluates affinity scores across the full pool of $N$ candidate experts:

\begin{equation}
\boldsymbol{\pi}_j = \text{Softmax}\!\bigl(W_r \, x_j + b_r\bigr) \in \mathbb{R}^{N},
\end{equation}

\noindent where $W_r \in \mathbb{R}^{N \times D}$ and $b_r \in \mathbb{R}^{N}$ are the trainable router parameters. The top-$k_j$ experts are then selected:

\begin{equation}
\hat{\boldsymbol{\pi}}_j^{(1:k_j)},\;\mathcal{S}_j^{(1:k_j)} = \text{TopK}\!\bigl(\boldsymbol{\pi}_j,\;k_j\bigr),
\end{equation}

\noindent where $\hat{\boldsymbol{\pi}}_j^{(1:k_j)}$ collects the $k_j$ largest affinity scores and $\mathcal{S}_j^{(1:k_j)}$ records the corresponding expert indices. The PA-MoE output for token $x_j$ is obtained via the re-normalized weighted fusion of the selected expert transformations:

\begin{equation}
\hat{x}_j = \sum_{m=1}^{k_j} \widetilde{\pi}_j^{(m)} \cdot E_{\mathcal{S}_j^{(m)}}(x_j), \qquad \widetilde{\pi}_j^{(m)} = \frac{\hat{\pi}_j^{(m)}}{\sum_{m'=1}^{k_j} \hat{\pi}_j^{(m')}},
\end{equation}

\noindent where $E_n(\cdot)$ represents the feed-forward transformation encapsulated by the $n$-th expert sub-network. The re-normalization step ensures that gating weights sum to unity, stabilizing gradient magnitudes during back-propagation.

\textbf{Remark.} Because heavy-precipitation tokens recruit more experts, PA-MoE implicitly constructs a richer functional basis for modeling the complex nonlinear dynamics of intense convection—mirroring the physical reality that extreme rainfall arises from the interaction of multiple atmospheric processes (moisture convergence, instability release, orographic forcing, etc.), each of which may be better captured by a dedicated specialist.

\subsection{Training Objectives and Optimization Strategy}

\subsubsection{Training Objectives}

{\bfseries Categorical Precipitation Loss.}
The decoder head converts the final hidden states $\mathbf{H}_{t+1}, \ldots, \mathbf{H}_{t+j}$ into discrete rainfall intensity categories through a stack of convolutional layers followed by non-linear activations, producing categorical forecasts $\hat{\mathbf{R}}_{t+1}^{\text{cat}}, \ldots, \hat{\mathbf{R}}_{t+j}^{\text{cat}}$. These predictions are spatially upsampled via learned transposed convolution to match the native grid resolution. To counteract the dominance of low-intensity classes inherent in the long-tailed rainfall distribution, we employ the soft Dice objective \cite{DiceLoss}, which measures volumetric overlap between predicted and ground-truth class masks and thereby naturally down-weights the contribution of over-represented categories. For a pixel at location $(i,j)$ with predicted class probability $\hat{p}_{i,j}^{(c)}$ and one-hot ground-truth label $y_{i,j}^{(c)}$ for category $c$, the per-sample Dice loss reads:
\begin{equation}
    \label{loss_pred}
    \mathcal{L}_{\text{pred}} = 1 - \frac{2\,\sum_{c=1}^{C_p} \hat{p}_{i,j}^{(c)}\, y_{i,j}^{(c)} + \epsilon}{\sum_{c=1}^{C_p} \bigl(\hat{p}_{i,j}^{(c)}\bigr)^2 + \sum_{c=1}^{C_p} y_{i,j}^{(c)} + \epsilon},
\end{equation}
where $C_p$ is the total number of precipitation categories and $\epsilon$ is a smoothing constant for numerical robustness.

{\bfseries PA-MoE Regularization Loss.}
The variable expert activation budget introduced by PA-MoE creates an additional need for routing regularization: without explicit constraints, the router may collapse onto a narrow subset of experts, leaving the remainder under-utilized. We address this through a composite auxiliary term comprising two complementary components.

\emph{(i) Dispatch Equity Penalty.} We quantify the dispersion of token-to-expert assignments across the batch. Let $U_e$ denote the total number of tokens dispatched to expert $e$ and $N_{\text{tok}} = B \cdot T \cdot H \cdot W$ the overall token count. The equity term minimizes the coefficient of variation of the usage vector:
\begin{equation}
\mathcal{L}_{\text{equity}} = \frac{\text{Std}\!\bigl(U_1/N_{\text{tok}},\;\ldots,\;U_N/N_{\text{tok}}\bigr)}{\text{Mean}\!\bigl(U_1/N_{\text{tok}},\;\ldots,\;U_N/N_{\text{tok}}\bigr) + \epsilon}.
\end{equation}

\emph{(ii) Router Decisiveness Penalty.} To prevent the gating distribution from degenerating into a near-uniform allocation that would negate the benefit of sparse expert selection, we penalize the Shannon entropy of the per-token routing distribution:
\begin{equation}
\mathcal{L}_{\text{decisive}} = \frac{1}{N_{\text{tok}}} \sum_{j=1}^{N_{\text{tok}}} \Bigl(-\sum_{e=1}^{N} \pi_{j,e}\,\log(\pi_{j,e} + \epsilon)\Bigr),
\end{equation}
where $\pi_{j,e}$ is the router probability assigned by token $j$ to expert $e$. Low entropy corresponds to sharp, confident routing decisions.

The full PA-MoE regularization loss merges both components:
\begin{equation}
\mathcal{L}_{\text{moe}} = \mathcal{L}_{\text{equity}} + \beta \cdot \mathcal{L}_{\text{decisive}},
\end{equation}
where $\beta$ governs the relative strength of the decisiveness term.

\subsubsection{Overall Optimization Objective}

The complete training loss aggregates the categorical prediction objective and the PA-MoE regularizer:
\begin{equation}
\mathcal{L}_{\text{total}} = \mathcal{L}_{\text{pred}} + \gamma \cdot \mathcal{L}_{\text{moe}},
\end{equation}
where $\gamma$ is a scalar coefficient that balances forecast accuracy against routing health. In practice, we set $\gamma$ to a small value (e.g., $10^{-2}$) so that the primary supervisory signal remains the precipitation prediction loss, while the auxiliary term exerts a gentle regularizing influence on expert utilization throughout training.

\subsubsection{Intensity-Aware Curriculum Strategy}

To further reinforce the model's sensitivity to tail-end rainfall events, we adopt a curriculum-style training schedule that progressively shifts supervisory emphasis toward higher precipitation intensities. In the initial training phase, all samples contribute equally to gradient updates, allowing the network to establish robust representations of the dominant dry and light-rain regimes. As training advances, we introduce an intensity-dependent sample weighting factor $w(r)$ that amplifies the gradient contribution of tokens with elevated rainfall:
\begin{equation}
w(r) = 1 + \eta \cdot \left(\frac{r}{r_{\text{ref}}}\right)^{\alpha_w},
\end{equation}
where $r$ is the local rainfall rate, $r_{\text{ref}}$ is a reference intensity (e.g., the 90th-percentile value from the training climatology), and $\eta, \alpha_w > 0$ are schedule hyperparameters that are linearly ramped up from zero during the first fraction of total training epochs. This graduated intensification prevents premature overfitting to rare extreme samples while ensuring that, by the conclusion of training, the model has devoted substantial learning capacity to the high-impact tail of the precipitation distribution.

\section{Experiments}

\subsection{Datasets}


A wide publicly available meteorological archive is employed to benchmark PA-Net: ERA5.

\textbf{ERA5.} Twelve hourly atmospheric fields serve as predictors: historical precipitation, relative humidity, air temperature, zonal wind (u), and meridional wind (v) at the 500\,hPa, 850\,hPa, and 1000\,hPa pressure levels. The native $0.25^{\circ}$ grid ($1440\times721$ cells) is cropped to a $221\times281$ regional domain spanning $140^{\circ}$E–$70^{\circ}$W and $0^{\circ}$–$55^{\circ}$N. Data from 2016–2018 form the training partition, 2019 is reserved for validation, and 2020 constitutes the held-out test set.



\subsection{Evaluation Protocol}

Precipitation intensity thresholds follow the categorization in Table~\ref{tab:data}. Two complementary metrics are reported.

\textbf{Intersection over Union (IoU)} quantifies the spatial overlap between forecast and observed rainfall masks:
\begin{equation}
\text{IoU} = \frac{|\mathcal{P} \cap \mathcal{G}|}{|\mathcal{P} \cup \mathcal{G}|},
\end{equation}
where $\mathcal{P}$ and $\mathcal{G}$ denote the sets of grid cells exceeding a given intensity threshold in the prediction and ground truth, respectively.

\textbf{Threat Score (TS)}, a standard operational verification measure, is derived from the contingency table entries—hits ($H$), false alarms ($F$), and misses ($M$)—defined relative to an intensity threshold $\theta$:
\begin{align}
H &= (\text{GT} \geq \theta) \;\wedge\; (\text{Pred} \geq \theta), \\
F &= (\text{GT} < \theta) \;\wedge\; (\text{Pred} \geq \theta), \\
M &= (\text{GT} \geq \theta) \;\wedge\; (\text{Pred} < \theta),
\end{align}
yielding:
\begin{equation}
\text{TS} = \frac{H}{H + M + F}.
\end{equation}

To illustrate with the heavy-rain category: IoU measures how well the predicted heavy-rain footprint coincides with the observed one, while TS evaluates detection skill by penalizing both missed events and false alerts. Following common practice, hourly forecasts are evaluated independently and six-hour averages are reported.

\subsection{Implementation Details}

\subsection{Quantitative Comparison}


\subsubsection{Results on ERA5}

All competing approaches receive identical input configurations to ensure an equitable comparison. Minor adaptations are applied to certain baselines (e.g., PFST) to accommodate the specific data format and prediction target of our task.

Table~\ref{tab:overallera5} summarizes the category-averaged results. PA-Net attains the best scores on both benchmarks. On ERA5, it surpasses the strongest competitor by at least 5.54 percentage points in IoU and 3.94 points in TS. Notably, PA-Net achieves these gains with only 18.90\,M parameters—an order of magnitude fewer than most baselines—demonstrating that the improvements stem from architectural design rather than brute-force capacity scaling.


\begin{table}[!tbp]
\centering
\caption{Category-averaged performance on the ERA5 dataset. Bold entries mark the highest scores.}
{
    \begin{tabular}{l|c|c|c|c}
        \toprule
        Model & Params(M) & MACs(G) & TS~$\uparrow$ & IoU~$\uparrow$ \\
        \midrule
        ConvLSTM  & 14.79 & 463.74 & 20.68 & 37.83 \\
        ConvGRU  & 17.19 & 604.8 & 20.41 & 38.03  \\
        TrajGRU  & {12.72} & 720.64 & 19.50 & 36.09  \\
        PredRNN  & 25.84 & 1103.64 & 17.95 & 36.58  \\
        PFST  & 40.65 & 472.97 & 18.95 & 38.05 \\
        SCCN  & 48.1 & 2990.48 & 19.77 & 38.01  \\
        SimVP  & 14.08 & 419.44 & 20.52 & 39.84   \\
        SimVP+gSTA & 10.48 & 330.96 & 20.27 & 39.46  \\
        TAU & 10.05 & 318.78 & 20.58 & 39.76 \\ 
        PA-Net & 18.90 & {426.94} & \textbf{24.62} & \textbf{45.38} \\
        \bottomrule
    \end{tabular}
} 
\label{tab:overallera5}
\end{table}

\subsection{Detailed Analysis}

\noindent \textbf{Breakdown by Precipitation Intensity.}
Table~\ref{tab:multi} disaggregates IoU on ERA5 across five intensity tiers. PA-Net consistently leads in every category, but the advantage is most striking in the tail regimes that PA-MoE is specifically designed to reinforce. For moderate and heavy rain, IoU improvements over the next-best baseline reach at least 6.12 and 3.35 percentage points, respectively. The rainstorm category exhibits the largest absolute gain—9.37 points—underscoring that the intensity-adaptive expert allocation effectively concentrates representational resources on the rarest and most challenging events. Importantly, the head-class (Rainless, Light Rain) scores remain competitive, confirming that enriching the tail does not come at the expense of dominant-regime accuracy.

\begin{table}[t]
	\centering	
	\setlength{\tabcolsep}{1mm}
    \caption{Per-category IoU (\%) on the ERA5 dataset stratified by rainfall intensity.}
	{
	\begin{tabular}{lccccc}
		\toprule
		Model & RL & LR & MR & HR & RS \\
		\midrule
		PFST  & 88.13 &  50.12 & 22.88 & 17.27 & 12.04 \\
		SCCN  &  87.13 &  51.42 & 20.16 & 17.5 & 13.85 \\
            SimVP & 87.28 & 50.97 & 25.89 & 21.02 & 14.04 \\
            SimVP+gSTA & 87.23 & 51.71 & 25.13 & 19.36 & 13.88 \\
            TAU & {88.41} & 52.30 & 26.33 & 19.31 & 12.58 \\
            PA-Net  & \textbf{88.89} & \textbf{57.78} & \textbf{32.45} & \textbf{24.37} & \textbf{23.41} \\
		\bottomrule
	\end{tabular} 
	}
\label{tab:multi}
\end{table}

\noindent \textbf{Temporal Evolution of Forecast Skill.}
Table~\ref{tab:overall_performances_ERA5} tracks hourly IoU and TS on ERA5 from 0–1\,h out to 5–6\,h. PA-Net achieves the highest scores at every lead time under both metrics. The advantage is especially pronounced at short horizons: at the 0–1\,h window, PA-Net exceeds the runner-up by 9.06 percentage points in IoU, indicating that the precipitation-adaptive routing captures rapidly evolving convective signals that competing methods fail to resolve. While all models exhibit the expected degradation with increasing lead time, PA-Net's decay curve is notably flatter, suggesting that the enriched expert ensemble mobilized for intense-precipitation tokens also benefits temporal coherence. Among the baselines, SCCN ranks second at 0–2\,h lead times—likely benefiting from its use of historical precipitation priors—but its skill collapses sharply beyond the two-hour mark, highlighting the fragility of approaches that lack explicit long-tail capacity management.

\begin{table}[t]\footnotesize
    \caption{Hourly IoU (\%) and TS (\%) on the ERA5 dataset across six one-hour forecast windows.}
	\renewcommand{\arraystretch}{0.8}
	\centering
    \setlength{\tabcolsep}{1mm}
    {
		\begin{tabular}{@{}c c c c c c c c}
                \toprule
                \multirow{3}*{Metric} & \multirow{3}*{Method} & \multicolumn{6}{c}{Prediction time (hours)}\\
                \cmidrule(lr){3-8} 
                & & 0$\sim$1 & 1$\sim$2 & 2$\sim$3 & 3$\sim$4 & 4$\sim$5 & 5$\sim$6\\
                \midrule
                \multirow{13}*{IoU$\uparrow$} 
                & Pers. & 31.10 & 28.64 & 26.98 & 25.78 & 24.89 & 24.22 \\
                & Clim. & 13.36 & 13.36 & 13.36 & 13.36 & 13.36 & 13.36 \\
                & W-Clim. & 16.68 & 16.68 & 16.68 & 16.68 & 16.68 & 16.68 \\
                & ConvLSTM & 40.52 & 40.10 & 38.84 & 37.39 & 35.85 & 34.29 \\
                & ConvGRU & 40.01 & 39.71 & 38.83 & 37.75 & 36.56 & 35.31  \\
                & TrajGRU & {40.57} & {40.12} & {36.45} & {35.02} & {32.72} & 31.63 \\
                & PredRNN  & 40.56 & 39.75 & 37.49 & 35.67 & 33.59 & 32.39 \\
                & PFST & 40.67 & 40.43 & 39.29 & 37.56 & 35.94 & 34.42 \\
                & SCCN & 44.72 & 44.70 & 38.69 & 35.30 & 33.28 & 31.37\\
                & SimVP & 43.91 & 43.01 & 41.27 & 39.04 & 36.87 & 34.91\\
                & SimVP+gSTA & 44.04 & 43.01 & 40.67 & 38.25 & 36.15 & 34.65\\
                & TAU & 43.73 & 42.73 & 40.93 & 38.88 & 36.96 & 35.58\\
                & PA-Net & \textbf{53.78} & \textbf{49.94} & \textbf{45.58} & \textbf{43.72} & \textbf{40.49} & \textbf{37.57} 
                \\
                \cmidrule(lr){1-8} 
                \multirow{13}*{TS$\uparrow$}
                & Pers. & 15.27 & 13.57 & 12.38 & 11.125 & 10.86 & 10.34 \\
                & Clim. & 6.36 & 6.36 & 6.36 & 6.36 & 6.36 & 6.36   \\
                & W-Clim. & 7.24 & 7.24 & 7.24 & 7.24 & 7.24 & 7.24  \\
                & ConvLSTM  & 22.48 & 22.15 & 21.32 & 20.39 & 19.40 & 18.37 \\
                & ConvGRU  & 21.89 & 21.63 & 20.98 & 20.20 & 19.35 & {18.45}  \\
                & TrajGRU  & 21.87 & 21.62 & 20.22 & 18.84 & 17.61 & 16.85 \\
                & PredRNN  & 20.13 & 19.75 & 18.62 & 17.48 & 16.34 & 15.43 \\
                & PFST  & 20.06 & 20.18 & 19.58 & 18.78 & 17.95 & 17.12 \\
                & SCCN & 26.33 & 24.17 & 19.62 & 17.39 & 16.11 & 14.98 \\
                & SimVP & 22.80 & 22.53 & 21.43 & 20.01 & 18.75 & 17.65\\
                & SimVP+gSTA & 22.73 & 22.14 & 20.93 & 19.67 & 18.52 & 17.61\\
                & TAU & 22.78 & 22.17 & 21.15 & 20.11 & 19.05 & 18.20 \\
                & PA-Net & \textbf{29.92} & \textbf{27.70} & \textbf{24.83} & \textbf{23.41} & \textbf{21.79} & \textbf{20.07} 
                \\
                \bottomrule
		\end{tabular}}
    \centering
    \label{tab:overall_performances_ERA5}
\end{table}



\begin{figure*}[t]
    \centering
    \includegraphics[width=\linewidth]{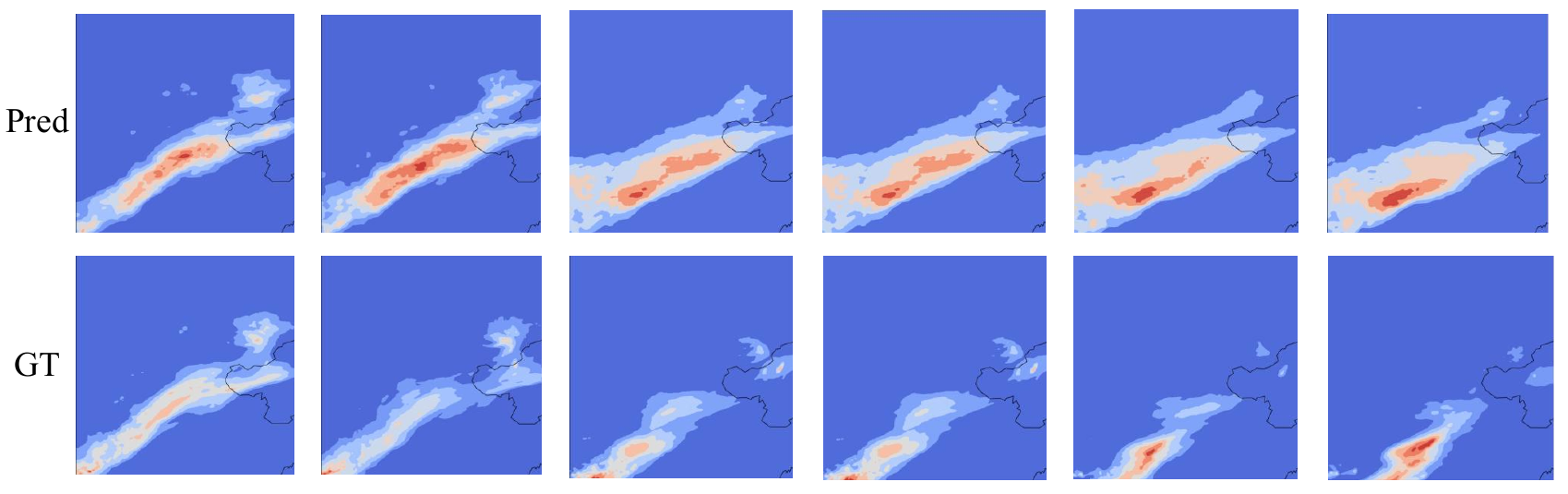}
    \caption{Six-hour precipitation forecast visualization on the ERA5 dataset. Columns from up to down: PA-Net and ground truth. Orange colors indicate heavier rainfall.}
    \label{fig:result}
\end{figure*}

\subsection{Ablation Study}

Table~\ref{tab:component} dissects the contribution of PA-Net's key components on the ERA5 benchmark. Starting from a vanilla Transformer baseline that lacks all proposed modules (40.93 IoU, 21.04 TS), we first introduce the Precipitation-Adaptive MoE (PA-MoE) in isolation. This single addition yields a substantial improvement of 2.15 points in IoU and 1.75 points in TS, reaching 43.08 and 22.79, respectively. This result confirms that intensity-driven expert routing provides a strong inductive bias: by channeling additional specialist sub-networks toward heavy-rainfall tokens, PA-MoE alone materially enhances the network's capacity to represent the challenging long-tail regime. When PA-MoE is further augmented with both Dual-Axis Compressed Latent Attention (DACLA) and the Intensity-Aware Training Protocol (IATP), performance climbs to the best configuration (45.38 IoU, 24.62 TS)—an additional gain of 2.30 points in IoU and 1.83 points in TS beyond PA-MoE alone. This incremental boost demonstrates clear synergistic effects: DACLA supplies richer spatiotemporal token representations to the adaptive router by efficiently capturing long-range dependencies along both spatial and temporal axes, while IATP progressively steers the optimization landscape toward the extreme-rainfall samples where PA-MoE deploys its largest expert ensembles. Overall, the full PA-Net configuration achieves a cumulative improvement of 4.45 points in IoU and 3.58 points in TS over the baseline, confirming that all three ingredients work in concert and are jointly essential for maximizing nowcasting skill.

\begin{table}[!tbp]
\centering
\caption{Component ablation on the ERA5 dataset. PA-MoE: Precipitation-Adaptive MoE; DACLA: Dual-Axis Compressed Latent Attention; IATP: Intensity-Aware Training Protocol.}
{
    \begin{tabular}{ccccc}
        \toprule
        PA-MoE & DACLA & IATP & IoU~$\uparrow$ & TS~$\uparrow$ \\
        \midrule
         &  &  &40.93&21.04\\
        \checkmark &  &  &43.08&22.79\\
        \checkmark &\checkmark & \checkmark &\textbf{45.38}&\textbf{24.62}\\
        \bottomrule
    \end{tabular}
} 
\label{tab:component}
\end{table}

\subsection{Qualitative Visualization}

Figure~\ref{fig:result} presents a representative case study comparing PA-Net on the ERA5 test set. The example depicts the six-hour precipitation outlook for East Asia initialized at 16:00 UTC on June 16, 2019, driven by the preceding six hours of multi-variate atmospheric input. Darker shading corresponds to higher rainfall accumulation. By routing additional experts to high-intensity tokens, PA-Net preserves fine-grained precipitation structure that fixed-capacity architectures tend to smooth out.


That said, all deep learning methods—including PA-Net—exhibit a tendency to produce spatially diffuse rainfall fields relative to ground truth, occasionally missing isolated, small-scale precipitation cells. Bridging this gap between continuous neural predictions and the inherently discrete, patchy nature of observed rainfall remains an open challenge for future work.


\section{Conclusion}

This paper presents the Precipitation-Adaptive Network (PA-Net), a Transformer-based framework that explicitly aligns model capacity with the long-tailed structure of rainfall intensity distributions. The core innovation, Precipitation-Adaptive MoE (PA-MoE), dynamically scales the number of activated experts per token according to local precipitation magnitude—mobilizing a broad specialist ensemble for rare heavy-rainfall events while processing abundant dry and light-rain tokens with a compact subset. This intensity-driven routing is complemented by Dual-Axis Compressed Latent Attention (DACLA), which factorizes spatiotemporal self-attention into sequential spatial and temporal stages with convolutional dimensionality reduction to accommodate million-scale token sequences, and an intensity-aware training protocol that progressively amplifies supervisory signals from extreme-rainfall samples. Comprehensive experiments on ERA5 demonstrate that PA-Net consistently outperforms existing baselines across all precipitation categories, with the most substantial improvements concentrated in the heavy-rain and rainstorm regimes. Despite these advances, the model still exhibits spatial over-smoothing for isolated convective cells, and the current tiered expert allocation relies on predefined climatological thresholds. Future work will explore continuous, learnable intensity-to-budget mappings and higher-resolution training to further close the gap with operational forecasting requirements.


\bibliographystyle{ACM-Reference-Format}
\bibliography{sample-base}


\begin{thebibliography}{24}


\ifx \showCODEN    \undefined \def \showCODEN     #1{\unskip}     \fi
\ifx \showDOI      \undefined \def \showDOI       #1{#1}\fi
\ifx \showISBNx    \undefined \def \showISBNx     #1{\unskip}     \fi
\ifx \showISBNxiii \undefined \def \showISBNxiii  #1{\unskip}     \fi
\ifx \showISSN     \undefined \def \showISSN      #1{\unskip}     \fi
\ifx \showLCCN     \undefined \def \showLCCN      #1{\unskip}     \fi
\ifx \shownote     \undefined \def \shownote      #1{#1}          \fi
\ifx \showarticletitle \undefined \def \showarticletitle #1{#1}   \fi
\ifx \showURL      \undefined \def \showURL       {\relax}        \fi
\providecommand\bibfield[2]{#2}
\providecommand\bibinfo[2]{#2}
\providecommand\natexlab[1]{#1}
\providecommand\showeprint[2][]{arXiv:#2}

\bibitem[Bauer et~al\mbox{.}(2015)]%
        {bauer2015quiet}
\bibfield{author}{\bibinfo{person}{Peter Bauer}, \bibinfo{person}{Alan Thorpe}, {and} \bibinfo{person}{Gilbert Brunet}.} \bibinfo{year}{2015}\natexlab{}.
\newblock \showarticletitle{The quiet revolution of numerical weather prediction}.
\newblock \bibinfo{journal}{\emph{Nature}} \bibinfo{volume}{525}, \bibinfo{number}{7567} (\bibinfo{year}{2015}), \bibinfo{pages}{47--55}.
\newblock


\bibitem[Bi et~al\mbox{.}(2023)]%
        {bi2023accurate}
\bibfield{author}{\bibinfo{person}{Kaifeng Bi}, \bibinfo{person}{Lingxi Xie}, \bibinfo{person}{Hengheng Zhang}, \bibinfo{person}{Xin Chen}, \bibinfo{person}{Xiaotao Gu}, {and} \bibinfo{person}{Qi Tian}.} \bibinfo{year}{2023}\natexlab{}.
\newblock \showarticletitle{Accurate medium-range global weather forecasting with 3D neural networks}.
\newblock \bibinfo{journal}{\emph{Nature}} (\bibinfo{year}{2023}), \bibinfo{pages}{1--6}.
\newblock


\bibitem[Dosovitskiy et~al\mbox{.}(2021)]%
        {DBLP:conf/iclr/DosovitskiyB0WZ21}
\bibfield{author}{\bibinfo{person}{Alexey Dosovitskiy}, \bibinfo{person}{Lucas Beyer}, \bibinfo{person}{Alexander Kolesnikov}, \bibinfo{person}{Dirk Weissenborn}, \bibinfo{person}{Xiaohua Zhai}, \bibinfo{person}{Thomas Unterthiner}, \bibinfo{person}{Mostafa Dehghani}, \bibinfo{person}{Matthias Minderer}, \bibinfo{person}{Georg Heigold}, \bibinfo{person}{Sylvain Gelly}, \bibinfo{person}{Jakob Uszkoreit}, {and} \bibinfo{person}{Neil Houlsby}.} \bibinfo{year}{2021}\natexlab{}.
\newblock \showarticletitle{An Image is Worth 16x16 Words: Transformers for Image Recognition at Scale}. In \bibinfo{booktitle}{\emph{International Conference on Learning Representations}}. \bibinfo{publisher}{OpenReview.net}.
\newblock


\bibitem[Gao et~al\mbox{.}(2024)]%
        {gao2024prediff}
\bibfield{author}{\bibinfo{person}{Zhihan Gao}, \bibinfo{person}{Xingjian Shi}, \bibinfo{person}{Boran Han}, \bibinfo{person}{Hao Wang}, \bibinfo{person}{Xiaoyong Jin}, \bibinfo{person}{Danielle Maddix}, \bibinfo{person}{Yi Zhu}, \bibinfo{person}{Mu Li}, {and} \bibinfo{person}{Yuyang~Bernie Wang}.} \bibinfo{year}{2024}\natexlab{}.
\newblock \showarticletitle{Prediff: Precipitation nowcasting with latent diffusion models}.
\newblock \bibinfo{journal}{\emph{Advances in Neural Information Processing Systems}}  \bibinfo{volume}{36} (\bibinfo{year}{2024}).
\newblock


\bibitem[Gao et~al\mbox{.}(2022a)]%
        {0001S0Z00Y22}
\bibfield{author}{\bibinfo{person}{Zhihan Gao}, \bibinfo{person}{Xingjian Shi}, \bibinfo{person}{Hao Wang}, \bibinfo{person}{Yi Zhu}, \bibinfo{person}{Yuyang Wang}, \bibinfo{person}{Mu Li}, {and} \bibinfo{person}{Dit{-}Yan Yeung}.} \bibinfo{year}{2022}\natexlab{a}.
\newblock \showarticletitle{Earthformer: Exploring Space-Time Transformers for Earth System Forecasting}. In \bibinfo{booktitle}{\emph{Advances in Neural Information Processing Systems 35: Annual Conference on Neural Information Processing Systems 2022, NeurIPS 2022, New Orleans, LA, USA, November 28 - December 9, 2022}}, \bibfield{editor}{\bibinfo{person}{Sanmi Koyejo}, \bibinfo{person}{S.~Mohamed}, \bibinfo{person}{A.~Agarwal}, \bibinfo{person}{Danielle Belgrave}, \bibinfo{person}{K.~Cho}, {and} \bibinfo{person}{A.~Oh}} (Eds.).
\newblock


\bibitem[Gao et~al\mbox{.}(2022b)]%
        {DBLP:conf/nips/0001S0Z00Y22}
\bibfield{author}{\bibinfo{person}{Zhihan Gao}, \bibinfo{person}{Xingjian Shi}, \bibinfo{person}{Hao Wang}, \bibinfo{person}{Yi Zhu}, \bibinfo{person}{Yuyang Wang}, \bibinfo{person}{Mu Li}, {and} \bibinfo{person}{Dit{-}Yan Yeung}.} \bibinfo{year}{2022}\natexlab{b}.
\newblock \showarticletitle{Earthformer: Exploring Space-Time Transformers for Earth System Forecasting}. In \bibinfo{booktitle}{\emph{Neural Information Processing Systems}}.
\newblock


\bibitem[Juanzhen et~al\mbox{.}(2014)]%
        {sun2014use}
\bibfield{author}{\bibinfo{person}{Sun Juanzhen}, \bibinfo{person}{Xue Ming}, \bibinfo{person}{Wilson James~W}, \bibinfo{person}{Zawadzki Isztar}, \bibinfo{person}{Ballard Sue~P}, \bibinfo{person}{Onvlee-Hooimeyer Jeanette}, \bibinfo{person}{Joe Paul}, \bibinfo{person}{Barker Dale~M}, \bibinfo{person}{Li Ping-Wah}, \bibinfo{person}{Golding Brian}, {et~al\mbox{.}}} \bibinfo{year}{2014}\natexlab{}.
\newblock \showarticletitle{Use of NWP for nowcasting convective precipitation: Recent progress and challenges}.
\newblock \bibinfo{journal}{\emph{Bulletin of the American Meteorological Society}} \bibinfo{volume}{95}, \bibinfo{number}{3} (\bibinfo{year}{2014}), \bibinfo{pages}{409--426}.
\newblock


\bibitem[Kenton and Toutanova(2019)]%
        {kenton2019bert}
\bibfield{author}{\bibinfo{person}{Jacob Devlin Ming-Wei~Chang Kenton} {and} \bibinfo{person}{Lee~Kristina Toutanova}.} \bibinfo{year}{2019}\natexlab{}.
\newblock \showarticletitle{Bert: Pre-training of deep bidirectional transformers for language understanding}. In \bibinfo{booktitle}{\emph{Proceedings of naacL-HLT}}, Vol.~\bibinfo{volume}{1}. Minneapolis, Minnesota.
\newblock


\bibitem[Kudo(2018)]%
        {kudo2018sentencepiece}
\bibfield{author}{\bibinfo{person}{T Kudo}.} \bibinfo{year}{2018}\natexlab{}.
\newblock \showarticletitle{Sentencepiece: A simple and language independent subword tokenizer and detokenizer for neural text processing}.
\newblock \bibinfo{journal}{\emph{arXiv preprint arXiv:1808.06226}} (\bibinfo{year}{2018}).
\newblock


\bibitem[Liu et~al\mbox{.}(2022)]%
        {liu2021swinv2}
\bibfield{author}{\bibinfo{person}{Ze Liu}, \bibinfo{person}{Han Hu}, \bibinfo{person}{Yutong Lin}, \bibinfo{person}{Zhuliang Yao}, \bibinfo{person}{Zhenda Xie}, \bibinfo{person}{Yixuan Wei}, \bibinfo{person}{Jia Ning}, \bibinfo{person}{Yue Cao}, \bibinfo{person}{Zheng Zhang}, \bibinfo{person}{Li Dong}, \bibinfo{person}{Furu Wei}, {and} \bibinfo{person}{Baining Guo}.} \bibinfo{year}{2022}\natexlab{}.
\newblock \showarticletitle{Swin Transformer V2: Scaling Up Capacity and Resolution}. In \bibinfo{booktitle}{\emph{International Conference on Computer Vision and Pattern Recognition (CVPR)}}.
\newblock


\bibitem[Liu et~al\mbox{.}(2021)]%
        {liu2021Swin}
\bibfield{author}{\bibinfo{person}{Ze Liu}, \bibinfo{person}{Yutong Lin}, \bibinfo{person}{Yue Cao}, \bibinfo{person}{Han Hu}, \bibinfo{person}{Yixuan Wei}, \bibinfo{person}{Zheng Zhang}, \bibinfo{person}{Stephen Lin}, {and} \bibinfo{person}{Baining Guo}.} \bibinfo{year}{2021}\natexlab{}.
\newblock \showarticletitle{Swin Transformer: Hierarchical Vision Transformer using Shifted Windows}. In \bibinfo{booktitle}{\emph{Proceedings of the IEEE/CVF International Conference on Computer Vision (ICCV)}}.
\newblock


\bibitem[Milletari et~al\mbox{.}(2016)]%
        {DiceLoss}
\bibfield{author}{\bibinfo{person}{Fausto Milletari}, \bibinfo{person}{Nassir Navab}, {and} \bibinfo{person}{Seyed{-}Ahmad Ahmadi}.} \bibinfo{year}{2016}\natexlab{}.
\newblock \showarticletitle{V-Net: Fully Convolutional Neural Networks for Volumetric Medical Image Segmentation}. In \bibinfo{booktitle}{\emph{3DV}}. \bibinfo{pages}{565--571}.
\newblock


\bibitem[Park and Kim(2022)]%
        {DBLP:conf/iclr/ParkK22}
\bibfield{author}{\bibinfo{person}{Namuk Park} {and} \bibinfo{person}{Songkuk Kim}.} \bibinfo{year}{2022}\natexlab{}.
\newblock \showarticletitle{How Do Vision Transformers Work?}. In \bibinfo{booktitle}{\emph{International Conference on Learning Representations}}. \bibinfo{publisher}{OpenReview.net}.
\newblock


\bibitem[Ramesh et~al\mbox{.}(2021)]%
        {ramesh2021zero}
\bibfield{author}{\bibinfo{person}{Aditya Ramesh}, \bibinfo{person}{Mikhail Pavlov}, \bibinfo{person}{Gabriel Goh}, \bibinfo{person}{Scott Gray}, \bibinfo{person}{Chelsea Voss}, \bibinfo{person}{Alec Radford}, \bibinfo{person}{Mark Chen}, {and} \bibinfo{person}{Ilya Sutskever}.} \bibinfo{year}{2021}\natexlab{}.
\newblock \showarticletitle{Zero-shot text-to-image generation}. In \bibinfo{booktitle}{\emph{International conference on machine learning}}. Pmlr, \bibinfo{pages}{8821--8831}.
\newblock


\bibitem[Sennrich(2015)]%
        {sennrich2015neural}
\bibfield{author}{\bibinfo{person}{Rico Sennrich}.} \bibinfo{year}{2015}\natexlab{}.
\newblock \showarticletitle{Neural machine translation of rare words with subword units}.
\newblock \bibinfo{journal}{\emph{arXiv preprint arXiv:1508.07909}} (\bibinfo{year}{2015}).
\newblock


\bibitem[Shi et~al\mbox{.}(2015)]%
        {ShiCWYWW15}
\bibfield{author}{\bibinfo{person}{Xingjian Shi}, \bibinfo{person}{Zhourong Chen}, \bibinfo{person}{Hao Wang}, \bibinfo{person}{Dit{-}Yan Yeung}, \bibinfo{person}{Wai{-}Kin Wong}, {and} \bibinfo{person}{Wang{-}chun Woo}.} \bibinfo{year}{2015}\natexlab{}.
\newblock \showarticletitle{Convolutional {LSTM} Network: {A} Machine Learning Approach for Precipitation Nowcasting}. In \bibinfo{booktitle}{\emph{Neural Information Processing Systems}}. \bibinfo{pages}{802--810}.
\newblock


\bibitem[Shi et~al\mbox{.}(2017)]%
        {PrecipitationNowcasting17}
\bibfield{author}{\bibinfo{person}{Xingjian Shi}, \bibinfo{person}{Zhihan Gao}, \bibinfo{person}{Leonard Lausen}, \bibinfo{person}{Hao Wang}, \bibinfo{person}{Dit{-}Yan Yeung}, \bibinfo{person}{Wai{-}Kin Wong}, {and} \bibinfo{person}{Wang{-}chun Woo}.} \bibinfo{year}{2017}\natexlab{}.
\newblock \showarticletitle{Deep Learning for Precipitation Nowcasting: {A} Benchmark and {A} New Model}. In \bibinfo{booktitle}{\emph{Neural Information Processing Systems}}. \bibinfo{publisher}{Curran Associates Inc.}, \bibinfo{pages}{5622--5632}.
\newblock


\bibitem[Suman et~al\mbox{.}(2021)]%
        {ravuri2021skilful}
\bibfield{author}{\bibinfo{person}{Ravuri Suman}, \bibinfo{person}{Lenc Karel}, \bibinfo{person}{Willson Matthew}, \bibinfo{person}{Kangin Dmitry}, \bibinfo{person}{Lam Remi}, \bibinfo{person}{Mirowski Piotr}, \bibinfo{person}{Fitzsimons Megan}, \bibinfo{person}{Athanassiadou Maria}, \bibinfo{person}{Kashem Sheleem}, \bibinfo{person}{Madge Sam}, {et~al\mbox{.}}} \bibinfo{year}{2021}\natexlab{}.
\newblock \showarticletitle{Skilful precipitation nowcasting using deep generative models of radar}.
\newblock \bibinfo{journal}{\emph{Nature}} \bibinfo{volume}{597}, \bibinfo{number}{7878} (\bibinfo{year}{2021}), \bibinfo{pages}{672--677}.
\newblock


\bibitem[Van Den~Oord et~al\mbox{.}(2017)]%
        {van2017neural}
\bibfield{author}{\bibinfo{person}{Aaron Van Den~Oord}, \bibinfo{person}{Oriol Vinyals}, {et~al\mbox{.}}} \bibinfo{year}{2017}\natexlab{}.
\newblock \showarticletitle{Neural discrete representation learning}.
\newblock \bibinfo{journal}{\emph{Advances in neural information processing systems}}  \bibinfo{volume}{30} (\bibinfo{year}{2017}).
\newblock


\bibitem[Wang et~al\mbox{.}(2017)]%
        {DBLP:conf/nips/WangLWGY17}
\bibfield{author}{\bibinfo{person}{Yunbo Wang}, \bibinfo{person}{Mingsheng Long}, \bibinfo{person}{Jianmin Wang}, \bibinfo{person}{Zhifeng Gao}, {and} \bibinfo{person}{Philip~S. Yu}.} \bibinfo{year}{2017}\natexlab{}.
\newblock \showarticletitle{PredRNN: Recurrent Neural Networks for Predictive Learning using Spatiotemporal LSTMs}. In \bibinfo{booktitle}{\emph{Neural Information Processing Systems}}. \bibinfo{pages}{879--888}.
\newblock


\bibitem[Wu et~al\mbox{.}(2021)]%
        {DBLP:conf/cvpr/WuY0L21}
\bibfield{author}{\bibinfo{person}{Haixu Wu}, \bibinfo{person}{Zhiyu Yao}, \bibinfo{person}{Jianmin Wang}, {and} \bibinfo{person}{Mingsheng Long}.} \bibinfo{year}{2021}\natexlab{}.
\newblock \showarticletitle{MotionRNN: {A} Flexible Model for Video Prediction With Spacetime-Varying Motions}. In \bibinfo{booktitle}{\emph{{IEEE} Conference on Computer Vision and Pattern Recognition}}. \bibinfo{publisher}{Computer Vision Foundation / {IEEE}}, \bibinfo{pages}{15435--15444}.
\newblock


\bibitem[Xiao et~al\mbox{.}(2022)]%
        {DBLP:conf/icdm/XiaoJMXP22}
\bibfield{author}{\bibinfo{person}{Xinyu Xiao}, \bibinfo{person}{Qizhao Jin}, \bibinfo{person}{Gaofeng Meng}, \bibinfo{person}{Shiming Xiang}, {and} \bibinfo{person}{Chunhong Pan}.} \bibinfo{year}{2022}\natexlab{}.
\newblock \showarticletitle{Spatiotemporal Contextual Consistency Network for Precipitation Nowcasting}. In \bibinfo{booktitle}{\emph{International Conference on Data Mining}}. \bibinfo{pages}{1257--1262}.
\newblock


\bibitem[Zhang et~al\mbox{.}(2023a)]%
        {zhang2023deep}
\bibfield{author}{\bibinfo{person}{Yifan Zhang}, \bibinfo{person}{Bingyi Kang}, \bibinfo{person}{Bryan Hooi}, \bibinfo{person}{Shuicheng Yan}, {and} \bibinfo{person}{Jiashi Feng}.} \bibinfo{year}{2023}\natexlab{a}.
\newblock \showarticletitle{Deep long-tailed learning: A survey}.
\newblock \bibinfo{journal}{\emph{IEEE Transactions on Pattern Analysis and Machine Intelligence}} \bibinfo{volume}{45}, \bibinfo{number}{9} (\bibinfo{year}{2023}), \bibinfo{pages}{10795--10816}.
\newblock


\bibitem[Zhang et~al\mbox{.}(2023b)]%
        {nowcastnet}
\bibfield{author}{\bibinfo{person}{Yuchen Zhang}, \bibinfo{person}{Mingsheng Long}, \bibinfo{person}{Kaiyuan Chen}, \bibinfo{person}{Lanxiang Xing}, \bibinfo{person}{Ronghua Jin}, \bibinfo{person}{Michael~I. Jordan}, {and} \bibinfo{person}{Jianmin Wang}.} \bibinfo{year}{2023}\natexlab{b}.
\newblock \showarticletitle{Skilful nowcasting of extreme precipitation with NowcastNet}.
\newblock \bibinfo{journal}{\emph{Nature}}  \bibinfo{volume}{619} (\bibinfo{year}{2023}), \bibinfo{pages}{526–532}.
\newblock


\end{thebibliography}

\end{document}